%% file: paper.tex
\documentclass[conference]{IEEEtran}
\usepackage{times}

\usepackage[numbers]{natbib}
\usepackage{multicol}
\usepackage[bookmarks=true]{hyperref}
\usepackage{amsmath}
\usepackage{amsfonts}
\usepackage{pifont}
\usepackage{bbm}
\usepackage{graphicx}
\usepackage[dvipsnames]{xcolor}
\hypersetup{
    colorlinks,
    linkcolor={red!80!black},
    citecolor={blue!80!black},
    urlcolor={blue!80!black}
}
\usepackage{booktabs}
\usepackage{subcaption}
\usepackage{multirow}
\usepackage{makecell}
\usepackage{xcolor}
\let\labelindent\relax
\usepackage{enumitem}
\setlist{noitemsep, topsep=2pt}

\usepackage{microtype}

\usepackage{listings}
\usepackage{xcolor}

\lstdefinestyle{constraintprompt}{
basicstyle=\ttfamily\small,
frame=single,
breaklines=true,
breakindent=0pt,
breakautoindent=false,
columns=fullflexible,
keepspaces=true,
showstringspaces=false,
}

\begin{document}

\title{Beyond Failure Recovery: An Engagement-Aware Human-in-the-loop Framework for Robotic Systems}

\author{
\IEEEauthorblockN{
Jiaying Fang, 
Joyce Yang, 
Zhanxin Wu, 
Bohan Yang, 
Tapomayukh Bhattacharjee
}

\IEEEauthorblockA{Cornell University}
}

\makeatletter
\let\@oldmaketitle\@maketitle
\renewcommand{\@maketitle}{\@oldmaketitle
\vspace{3pt}
  \begin{center}
  \captionsetup{type=figure}
  \includegraphics[width=\textwidth]{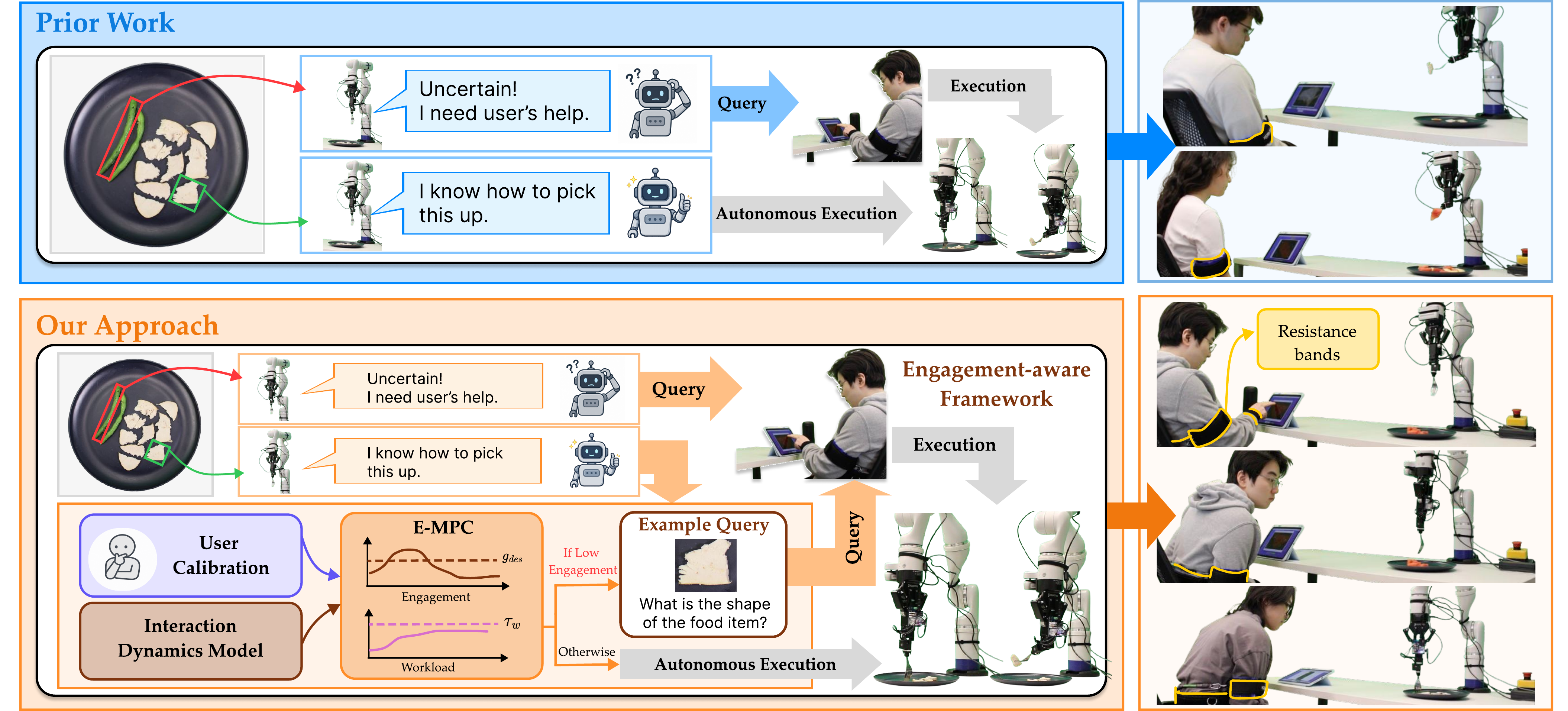}
    \captionof{figure}{\small{We propose \textbf{Engagement-aware MPC (E-MPC)}, a human-in-the-loop framework for robotic systems that explicitly models user engagement. E-MPC involves users in the robot's decision-making not only when user's assistance is needed, but also when interaction is desired to sustain user's preferred engagement level throughout the task.}}
    \label{fig:introduction}
    \vspace{-20pt}
    \end{center}
}
\makeatother

\maketitle
\addtocounter{figure}{-1}

\begin{abstract}
\input{01_abstract}
\end{abstract}

\IEEEpeerreviewmaketitle

\section{Introduction}
\input{02_introduction}

\section{Related Works}
\input{03_related_works}

\section{Problem Formulation}
\input{04_problem_formulation}

\section{Approach}
\input{05_approach}

\section{Evaluation Scenario}
\input{06_application}

\section{Simulation Experiments}
\input{07_results}

\section{Real-robot User Study}
\input{08_user_study}

\section{Discussions and Limitations}
\input{09_discussions}

\section*{Acknowledgments}
This work was partly funded by National Science Foundation IIS \#2132846, and CAREER \#2238792. The authors would like to additionally thank Rohan Banerjee for helping with the workload model, Yunting Yan for helping with the user study design, and all of the participants in our user study.


\bibliographystyle{plainnat}
\bibliography{references}

\clearpage
\setlength{\textfloatsep}{6pt}
\setlength{\floatsep}{6pt}
\setlength{\dbltextfloatsep}{6pt}
\setlength{\dblfloatsep}{6pt}
\setlength{\abovecaptionskip}{3pt}
\setlength{\belowcaptionskip}{0pt}
\input{10_appendix}
\end{document}

%% file: 01_abstract.tex
Conventional human-in-the-loop approaches typically involve users only when a robot encounters failure or uncertainty, treating humans primarily as tools to improve robot performance. However, in many human-centered robotics settings, interaction should support user engagement, keeping users meaningfully involved in decision-making rather than limiting them to failure-driven interventions. For many users, this cannot be achieved through limited, failure-driven interaction alone. This is particularly compelling in physical caregiving, where mobility limitations can reduce users' ability to intervene or modulate the robot's behavior in the moment. As a result, interaction policies that engage users only upon failure may further reduce engagement by relegating users to passive observers for long stretches of the task. For example, a user with mobility limitations may experience reduced engagement when being continuously and passively fed by a robot. At the same time, overly frequent interaction can be tiring and increase the user's workload.

To address this trade-off, we propose Engagement-aware MPC (E-MPC), a user-engagement-aware method that plans interaction to maintain engagement while respecting a workload constraint. E-MPC leverages a user interaction dynamics model that captures how user engagement evolves as a function of both the frequency and type of interaction. Rather than requesting input only when difficulties arise during task execution, the robot proactively considers the user's preferred level of engagement throughout the task, balancing autonomy and interaction while ensuring task success. We evaluate E-MPC in simulation with several ablations and baseline comparisons. Baselines optimize for task success alone or jointly for user workload and task success. Results demonstrate the effectiveness of our approach across diverse user personas. In addition, we conduct a real-world user study with participants with emulated mobility limitations on a robot-assisted bite acquisition system, showing that E-MPC improves user experience while maintaining task success. Website: \href{https://emprise.cs.cornell.edu/empc}{https://emprise.cs.cornell.edu/empc}

%% file: 02_introduction.tex
Human-in-the-loop robotic policies has traditionally been motivated by the goal of improving the robustness and reliability of autonomous robot policies. In such settings, human guidance can help provide feedback \cite{shi2024yell, sharma2022correcting, zha2024distilling} or assist with difficult perception and control challenges \cite{ross2011reduction, kelly2019hg, liu2025robot}. As robots become more capable, these human-in-the-loop mechanisms are increasingly being deployed in human-centered applications like caregiving and household robots, where failures can be costly and reliable execution is essential.

Despite these advances, current interaction strategies are often designed from a robot-centric perspective: the human is consulted mainly when the robot requires additional information to complete an action successfully. In many household or assistive tasks where the human is involved, the quality of robot's assistance is shaped not only by task success and failure recovery, but also by how the person experiences the process \cite{bhattacharjee2020more, collier2025sense}. This is particularly critical in caregiving applications, where maintaining engagement helps preserve a sense of agency and long-term acceptance of robotic assistance. For example, during robot-assisted feeding, users may prefer to shape how the robot picks up and delivers food, rather than passively receiving continuous robot autonomy \cite{bhattacharjee2020more}. However, meaningful interaction must be balanced carefully: interaction that is too sparse may leave users disengaged, while interaction that is too frequent can increase user workload \cite{banerjee2025ask}. This tension is especially pronounced in long-horizon tasks, where the robot repeatedly makes decisions over time and the user’s preferences for engagement vary across individuals and circumstances.

In this work, we propose \textbf{Engagement-aware Model Predictive Control} (\textit{E-MPC}), a framework that explicitly treats user engagement as an objective during task execution. Rather than viewing interaction solely as a mechanism to resolve uncertainty, \textit{E-MPC} enables the robot to proactively decide when and how to involve the user in order to maintain a desired level of engagement throughout the task. Our approach leverages a predictive model of interaction dynamics, capturing how engagement evolves based on both the frequency and type of interaction, while also enforcing constraints on user workload and task success. \textit{E-MPC} provides a principled way to balance autonomy and interaction in long-horizon tasks. In general, our contributions are:
\begin{itemize}
    \item A human-in-the-loop framework for robotic systems that jointly reasons about user engagement, user workload and task success to improve user experience. 
    \item An interaction dynamics model that models user engagement as a function of interaction frequency and type.
    \item Extensive evaluations of the framework both in simulation with different personas and on a real robot-assisted bite acquisition system through an in-lab user study with participants with emulated mobility limitations.
\end{itemize}

%% file: 03_related_works.tex
\subsection{Engagement in Human-Robot Interaction}
Prior research in Human-Robot Interaction has highlighted the importance of sustaining user engagement in long-horizon interactions \cite{oertel2020engagement, laban2022user, del2020you, sorrentino2024definition}. Studies suggest that engagement can diminish over time without deliberate interaction design, even when robotic task execution remains successful \cite{del2020you, 8673076, matheus2025long}. Engagement has also been linked to a user’s sense of agency \cite{collier2025sense, yang2025preserving, glawe2025human} - the perceived ability to influence ongoing outcomes - which has been shown to support psychological well-being \cite{ryan2000self} and acceptance of assistive technologies \cite{cornelio2022sense, nertinger2024acceptance}. In addition, in assistive robotics, users can prefer meaningful involvement in robot decision-making over full autonomy \cite{nasir2022if}, valuing a sense of control even when it may reduce task efficiency \cite{bhattacharjee2020more}. 

A related line of work focuses on recognizing and detecting engagement during interaction \cite{rich2010recognizing, moon2014meet, sanghvi2011automatic}. However, these approaches do not model engagement as an objective that can be actively shaped in long-horizon tasks. More closely related to our work, Del Duchetto and Hanheide~\cite{del2022learning} consider engagement in interaction planning, but aim to maximize engagement inferred from sensory input. In contrast, we regulate engagement toward a user-specified target and model it as a continuous state governed by engagement dynamics. Our work thus introduces a principled engagement-aware interaction planning framework that treats engagement as a controllable objective.

\subsection{Human-in-the-loop Robotic Policies}
Most human-in-the-loop robotic policies seek human assistance when the robot is uncertain about how to complete the task or encounters execution failure \cite{shi2024yell, kelly2019hg, ross2011reduction, liu2025robot, sharma2022correcting, zha2024distilling, kim2025srt, ren2023robots, li2023interactive, ye2024morpheus}. In the DAgger line of work \cite{ross2011reduction, kelly2019hg, liu2025robot}, the human expert is iteratively queried for good actions on challenging or out-of-distribution observations. In \cite{sharma2022correcting, kim2025srt, ren2023robots}, users provide corrective feedback to the robot policy to help it complete the task. Recently, some work extends the human-in-the-loop framework to jointly optimize task success and human workload, recognizing that excessive querying can increase users' workload \cite{banerjee2025ask, banerjee2026modularhil}. However, these methods do not explicitly incorporate user engagement objectives into policy optimization. In contrast, our method models and optimizes for user engagement in an MPC framework.

Another line of human-in-the-loop robotic policies is shared autonomy, which blends continuous human inputs with autonomous robot control \cite{Javdani-RSS-15, dragan2013policy}. While these approaches operate at the control level, our setting instead considers a higher-level interaction planner that selects from a discrete set of query actions.

%% file: 04_problem_formulation.tex
We investigate the problem of achieving successful task completion with the robotic systems while ensuring bounded user workload and optimizing user engagement toward a desired engagement level. The robotic systems considered in this work are designed to accomplish specific tasks with a human user involved. At each time step, the robot may either query the user for input or proceed autonomously.

We assume access to a skill library $\mathcal{S} = \{\phi^1, \phi^2, \dots, \phi^N\}$ consisting of $N$ skills that the robot can use to complete the task. Task execution involves a planner that produces a sequence of skills over discrete time steps $t=1, 2, \dots, T$ to achieve the goal. Each skill $\phi^i_t$ executed at time $t$ is associated with a confidence value $
c_t = \mathcal{C}(\phi^i_t, \mathbf{o}_t) \in [0,1]$,
where $\mathbf{o}_t$ denotes the observation at time $t$ and $\mathcal{C}$ is the confidence estimation function. Larger values of $c_t$ indicate a higher likelihood of successful skill execution. If a skill execution fails, the system may retry up to $\tau_r$ times. 

In our human-in-the-loop robotic system, the robot may interact with the user by selecting an interaction action $\mathbf{q}_t \in \mathcal{Q}$,
where $\mathcal{Q}$ denotes the finite set of available query types. Because the robot may also choose to proceed autonomously without querying the user, this set includes a no-query action represented by the zero vector $\mathbf{0}=[0,0]\in\mathcal{Q}$. Each query action is parameterized by a two-dimensional vector $
\mathbf{q}_t = [q_{d,t},\, q_{rd,t}]$,
where $q_{d,t}$ denotes the query difficulty and $q_{rd,t}$ denotes the query response difficulty.

Each skill $\phi^i$ is associated with a set of \emph{task-assistive} query types,
$\mathcal{Q}^{\mathrm{assist},\phi^i} \subseteq \mathcal{Q}$,
which provides functional user input that can directly help the robot successfully executing the skill under uncertainty. For example, in a bite-acquisition system, when the autonomous policy is highly uncertain about where to skewer a food item, a drawing-based query that asks the user to indicate the skewering point on an image can directly support the successful execution of the skewering skill. In addition, each skill $\phi^i$ is associated with a set of \emph{feasible} query types, $\mathcal{Q}^{\mathrm{feas},\phi^i} \subseteq \mathcal{Q}$,
representing all task-relevant queries that the robot may ask during execution. This feasible set includes the \emph{task-assistive} queries, $\mathcal{Q}^{\mathrm{assist},\phi^i} \subseteq \mathcal{Q}^{\mathrm{feas},\phi^i}$,
as well as queries that may increase user engagement but do not really help the robot complete the task. For instance, the robot may ask ``fake'' queries, such as asking the user about the shape of a food item. These interactions can enhance engagement but do not affect execution of the skewering skill. At each time step $t$, before executing skill $\phi_t^i$, the robot can either ask a query by selecting $\mathbf{q}_t \in \mathcal{Q}^{\mathrm{feas},\phi_t^i}$ or execute the skill autonomously ($\mathbf{q}_t=\mathbf{0}$). If a query is asked, the user provides a response, and the robot may incorporate this information to carry out the skill. We denote the time steps at which the robot issues a query ($\mathbf{q}_t \neq \mathbf{0}$) by $\{t_k\}_{k=1}^K$, where $t_k \in \{1,\dots,T\}$, $K \leq T$.

Finally, we represent user workload at time $t$ as $w_t \in [0,1]$, and user engagement as $g_t \in [0,1]$. The user's desired engagement level is denoted by $g_{\mathrm{des}} \in [0,1]$.

%% file: 05_approach.tex
\begin{figure*}[!t]
    \centering
    \includegraphics[width=\linewidth]{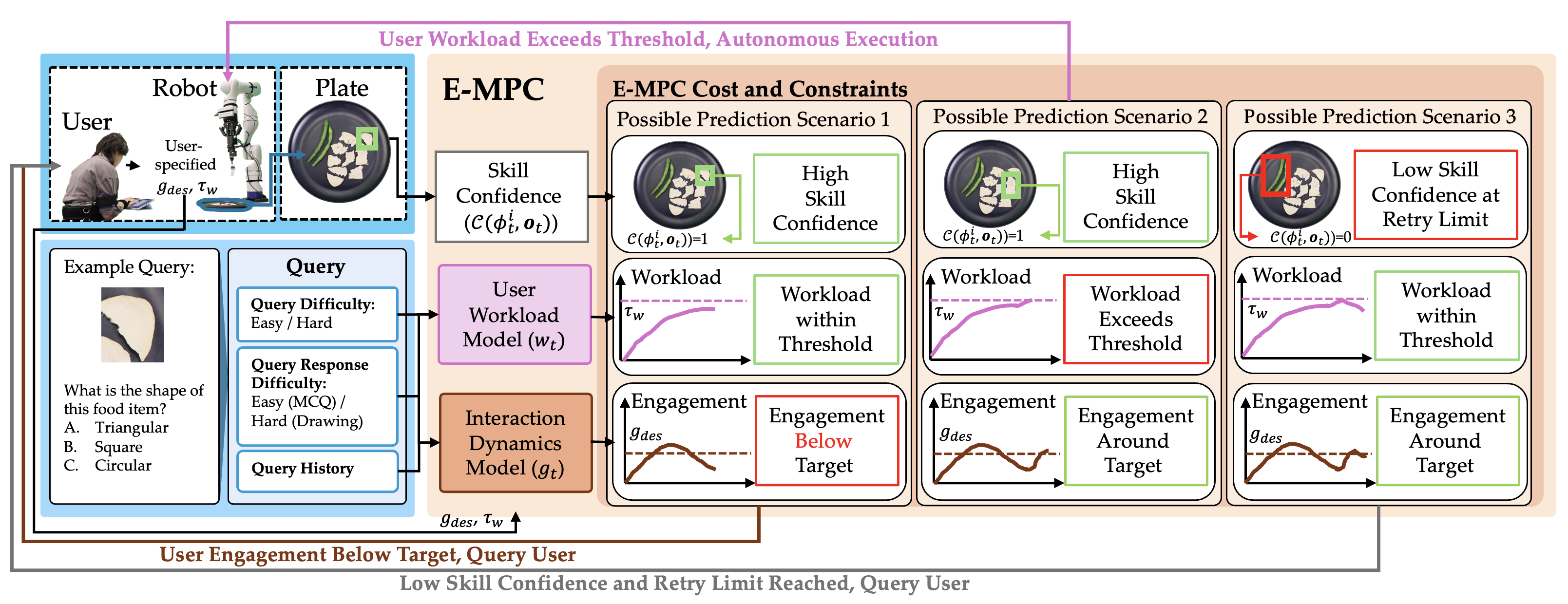}
    \caption{\small{\textbf{E-MPC framework}. \textit{E-MPC} proactively issues user queries to regulate engagement $g_t$ toward a user-specified desired level $g_{des}$ throughout task execution. By predicting the engagement effect of different query types, \textit{E-MPC} selects interactions that provide engagement boosts while keeping user workload $w_t$ bounded. When task success is at risk (low confidence $\mathcal{C}(\phi^i_t, \mathbf{o}_t)$ and retries exhausted), the controller requires \emph{task-assistive} queries, and replans in a receding-horizon manner.}}
    \label{fig:pipeline}
    \vspace{-20pt}
\end{figure*}
\subsection{Interaction Dynamics Modeling}
\label{sec:engagement_model}
\subsubsection{Engagement-Related Interaction Variables} We hypothesize that user engagement is related to: 
\begin{itemize}
    \item \textbf{Query response difficulty $q_{rd,t}$}: This represents the different response difficulties of a robot query. Different query response difficulties can lead to different engagement outcomes. We consider two types of query response difficulties: (a) the response difficulty of a multiple-choice question, for example, asking the user to choose one skill from multiple skills (b) the response difficulty of a drawing-based query, for example, asking user to draw action points or a bounding box on an image.
    \item \textbf{Query frequency $\delta_t$}: This represents the time gap between queries. Specifically, $\delta_t = \mathrm{seconds\_per\_timestep} \times (t - t_k)$, represents the time gap in seconds between the current time step $t$ and the time step $t_k$ which the previous query happened.
\end{itemize}

\subsubsection{Interaction Dynamics Model} Inspired by the above two variables, we treat engagement $g_t$ as a latent internal user state that is not directly observable, but estimated online through interaction outcomes and calibrated query impacts. We model the interaction dynamics as:
\begin{equation}
    g_t=e^{-\lambda_g\delta_t} g_{t_k} + (1 - e^{-\lambda_g\delta_t}) k_{q_{rd}}q_{rd,t},
    \label{eq:engagement_dynamics_model}
\end{equation}
where
\begin{equation}
q_{rd,t}=\mathbf{w}^{T}_{\mathrm{resp}} \mathbf{q}_t,
\end{equation}
and $\lambda_g > 0$ is the decay rate. $g_{t_k}$ denotes the engagement level at the most recent timestep $t_k$ when a query action was issued. The operator $\mathbf{w}_{\mathrm{resp}}=[0, 1]$ projects the interaction action $\mathbf{q}_t = [q_{d,t}, q_{rd,t}] \in \mathcal{Q}$ onto the query response difficulty component $q_{rd,t}$. $k_{q_{rd}}$ denotes an one-to-one mapping from query response difficulty $q_{rd,t}$ to query impact values in $[0, 1]$, which capture the immediate engagement boost induced at the moment of interaction. In real-robot experiments, this mapping is obtained through a user-specific calibration process conducted before the user study. When no query is issued at time $t$, i.e., $\mathbf{q}_t = \mathbf{0}$, the immediate engagement boost $k_{q_{rd}}q_{rd,t}$ evaluates to zero, and engagement evolves purely via decay. 

The proposed interaction dynamics model exhibits several key behavioral characteristics:
\begin{itemize}
    \item Engagement evolves as a combination of discounted prior engagement level and instantaneous engagement impulse. The effect of query frequency $\delta_t$ is captured through the discounting of prior engagement. The effect of query response difficulty $q_{rd,t}$ is encoded in the instantaneous engagement impulse.  
    \item Older interactions gradually fade, while recent ones exert greater influence, reflecting the temporally evolving nature of engagement \cite{sun2017sensing, o2008user}.
    \item Different query response difficulties lead to different query impacts, which aligns with the finding in \cite{bijkerk2023measuring, doherty2018engagement}. The engagement level is bounded by the range of the query impacts.  
\end{itemize}

We can summarize the parameter $\lambda_g$ has the following roles: It controls the relative weighting between previous engagement level and the instantaneous engagement impulse. A larger $\lambda_g$ means less persistent engagement memory and more sensitive to current engagement impulse. A smaller $\lambda_g$ means more persistent engagement memory and less sensitive to current engagement impulse. 

\subsection{User Workload Modeling}
\label{sec:workload_model}
Following \cite{banerjee2025ask}, we directly adopt a linear discrete-time user workload model: 
\begin{equation}
    w_t = \gamma w_{\mathrm{init}} + \sum_{j=0}^{H-1}{\mathbf{w}_j}^{T}\mathbf{q}_{t-j} + w_0
    \label{eq:workload}
\end{equation}

where $\gamma w_{\mathrm{init}}$ is the weighted initial workload, $\mathbf{q}_{t-j} = [q_{d, t-j}, q_{rd,t-j}]$ is the interaction action happened $j$ time steps in the past, $\mathbf{w}_j$ represents the effect of the interaction action $\mathbf{q}_{t-j} \in \mathcal{Q}$ on the workload at the current time $t$, $H$ is the length of the interaction history the workload model considers, and $w_0$ is a bias term.

\subsection{Engagement-aware MPC}
Given a robotic system, we aim to ensure task success while bounding user workload and optimizing user engagement toward a user-specified desired engagement level. This problem is challenging due to the subjectiveness in engagement and workload modeling, as well as the noise in the user’s stated engagement preference. We address these challenges by formulating the problem as a model predictive control (MPC) framework that jointly reasons about task success, user workload, and engagement level, while allowing the user to correct their engagement preference $g_{des}$ online. The MPC optimizes over interaction actions $\mathbf{q}_t \in \mathcal{Q}$ only; task-level skills are provided by an external planner.

\subsubsection{State Variables and Control Input} To reason about user engagement and user workload, the controller maintains three state variables:
\[
    g_t \in [0,1],
    \qquad
    g_{des} \in [0,1],
    \qquad
    w_t \in [0,1].
\]
$g_t$ captures the current estimated user engagement level at time $t$, $g_{des}$ is the desired user engagement level specified by the user. We assume that $g_{des}$ remains fixed over each MPC rollout, but may be updated between rollouts based on user feedback. $w_t$ is the estimated user workload. The dynamics model of $g_t$ and $w_t$ are explained in Sec. \ref{sec:engagement_model} and Sec. \ref{sec:workload_model}. The control of the MPC controller is the interaction action $\mathbf{q}_t \in \mathcal{Q}$.

\subsubsection{User Desired Engagement Level} The user desired engagement level $g_{des}$ is a self-reported preference, which specifies how actively the user wishes to participate in decision-making during the interaction and serves as the target engagement level $g_{des}$ for the system. Users can specify $g_{des}$ prior to the robot system rollout and update it during the rollout.

\subsubsection{Engagement Tracking Cost} At prediction step $l$, let $g_{t+l}$ be the predicted engagement state. The MPC cost is
\begin{equation}
    \begin{aligned}
    \mathcal{J}_{g}
    &= (g_{des} - g_{t+l})^2
    \end{aligned}
    \label{eq:stage_cost_hybrid}
\end{equation}
This term encourages the estimated engagement level $g_{t+l}$ to track the desired engagement level $g_{des}$.

\subsubsection{Task Success Constraint}
A task-related constraint is set to ensure the successful completion of the task. At prediction step $l$, the confidence of the skill to be executed is $\mathcal{C}(\phi^i_{t+l}, \mathbf{o}_{t+l})$. When this confidence falls below a threshold $\tau_c$, and the current number of retries $r^{\phi^i_{t+l}}_t$ of the skill reaches the maximum allowed number of retries $\tau_r$, the robot is required to seek user assistance. Otherwise, the robot may either select any query from the feasible set $\mathcal{Q}^{feas,\phi^i_t}$ or proceed autonomously without querying the user. Accordingly, we impose the
following constraint on the interaction action:
\begin{equation}
\mathbf{q}_{t+l}
\in
\begin{cases}
\mathcal{Q}^{\mathrm{assist}, \phi^i_{t+l}}, & \text{if  } \mathcal{C}(\phi^i_{t+l}, \mathbf{o}_{t+l}) \leq \tau_{c} \\ &\text{ and } r^{\phi^i_{t+l}}_t = \tau_r, \\
\mathcal{Q}^{\mathrm{feas}, \phi^i_{t+l}}\cup \{\mathbf{0}\} \, & \text{otherwise}.
\end{cases}
\label{eq:consequential_constraint}
\end{equation}

\subsubsection{User Workload Constraint} To avoid making the user feel overwhelmed by the robot queries, we also set a safe workload threshold $\tau_{w}$. At prediction step $l$, let 
$w_{t+l}$ be the predicted workload, the workload constraint is formulated as a cost term:
\begin{equation}
    \mathcal{J}_w = \lambda_w \big[\max(0,\, w_{t+l} - \tau_w)\big]^2,
    \label{eq:workload_constraint}
\end{equation}
where $\lambda_w$ is a large constant. This term allows the robot to exceed the workload threshold $\tau_w$ if necessary, but any violation is heavily penalized by the quadratic cost.

\subsubsection{Finite-Horizon MPC Optimization} At time $t$, with prediction horizon $L$, the controller solves:
\begin{equation}
\begin{aligned}
\min_{\{\mathbf{q}_{t+l}\}} \quad &
\sum_{l=0}^{L-1}
\Big(
(g_{\mathrm{des}} - g_{t+l})^2
+ \lambda_w \, \big[\max(0,\, w_{t+l} - \tau_w)\big]^2
\Big)
\\
\text{s.t.}\quad & \text{for } l = 0,\ldots,L-1, \\
& 
\mathbf{q}_{t+l}
\in
\begin{cases}
\mathcal{Q}^{\mathrm{assist}, \phi^i_{t+l}}, & \text{if  } \mathcal{C}(\phi^i_{t+l}, \mathbf{o}_{t+l}) \leq \tau_{c} \\ &\text{ and } r^{\phi^i_{t+l}}_t = \tau_r, \\
\mathcal{Q}^{\mathrm{feas}, \phi^i_{t+l}}\cup \{\mathbf{0}\} \, & \text{otherwise}.
\end{cases}
\end{aligned}
\label{eq:mpc_problem_hybrid}
\end{equation}

Only the first control input is applied, after which the horizon recedes and the optimization is repeated.

%% file: 06_application.tex
We now instantiate the proposed \textit{E-MPC} framework on a robot-assisted bite acquisition system. Robot-assisted feeding enhances the autonomy of individuals with mobility limitations \cite{jenamani2025feast, gordon2024adaptable, jenamani2024feel, bhattacharjee2020more}. Respecting and supporting user engagement in long-horizon tasks like bite acquisition in feeding can further empower users and promote a sense of agency \cite{jenamani2024flair, jenamani2025feast}.

\subsection{System Overview} The goal of our robot-assisted bite acquisition system is to acquire food items from a plate using a utensil. At each step of the task plan, the robot locates a food item from an image of the plate and executes one food acquisition skill. If the skill execution fails, the system can retry two times ($\tau_r=2$) on the same food item after the first failure. Following \cite{banerjee2026modularhil}, we estimate the acquisition 6D pose for different food items using a skill parameter estimation model, which a transformer-based vision-language-action model. The model takes as input a textual skill description, along with the current bounding box image as visual context, and outputs the corresponding acquisition pose. The primary uncertainty in this system comes from this skill parameter estimation model. To get human assistance, the robot can ask the user to draw the acquisition position and angle on the bounding box image.

\subsection{Skill Library}
The skill library $\mathcal{S}_{\mathrm{food}}$ of the robot-assisted bite acquisition system consists of three acquisition skills:
\begin{itemize}
    \item \textit{Skewering} ($\phi^{1}$): Skewers the target food item. The system estimates the 3D interaction point between the utensil and the food, along with the utensil roll angle.
    \item \textit{Scooping} ($\phi^{2}$): Scoops the food item. The system estimates the 3D start and end points of the scooping trajectory.
    \item \textit{Twirling} ($\phi^{3}$): Twirls the food item. The system estimates the 3D interaction point and the corresponding utensil roll angle required for twirling.
\end{itemize}

\subsection{Skill Confidences and Query Types}
\label{sec:query_types}
The confidence score for each acquisition skill is modeled as a binary value, derived from two sources: (1) the uncertainty of the transformer-based skill parameter estimation model, and (2) a failure identification module. The uncertainty of the skill parameter estimation model comes from the variance of the its predictions using Monte Carlo Dropout \cite{banerjee2026modularhil}. The failure detection module detects whether the predicted interaction point (e.g., a skewering location) is within the segmentation mask of the food item (see Appendix for details). Since food acquisition failures often stem from inaccurate parameter estimation, we design the following \emph{task-assistive queries} to obtain user assistance:

\begin{itemize}
    \item \textit{Skewering} query: The user is shown a cropped bounding-box image of the target food item and asked to draw the skewer point and angle.
    \item \textit{Scooping} query: The user is shown a cropped bounding-box image and asked to draw the start and end points of the scooping trajectory.
    \item \textit{Twirling} query: The user is shown a cropped bounding-box image and asked to draw the twirling point and angle.
\end{itemize}

In addition to these required queries, we introduce an auxiliary query type intended solely to promote user engagement:

\begin{itemize}
    \item \textit{Food item shape} query: The user is shown a cropped image of the food item and asked to select its shape from multiple-choice options.
\end{itemize}

The query type of \emph{task-assistive} queries is denoted as
$\mathbf{q}_t^{\mathrm{food,points}} = [q_{d,t}=2(\text{Hard}),\, q_{rd,t}=2(\text{Hard})]$,
while the engagement-only query type is denoted as
$\mathbf{q}_t^{\mathrm{food,shape}} = [q_{d,t}=1(\text{Easy}),\, q_{rd,t}=1(\text{Easy})]$.

%% file: 07_results.tex
\begin{figure*}[!t]
    \centering
    \includegraphics[width=\linewidth]{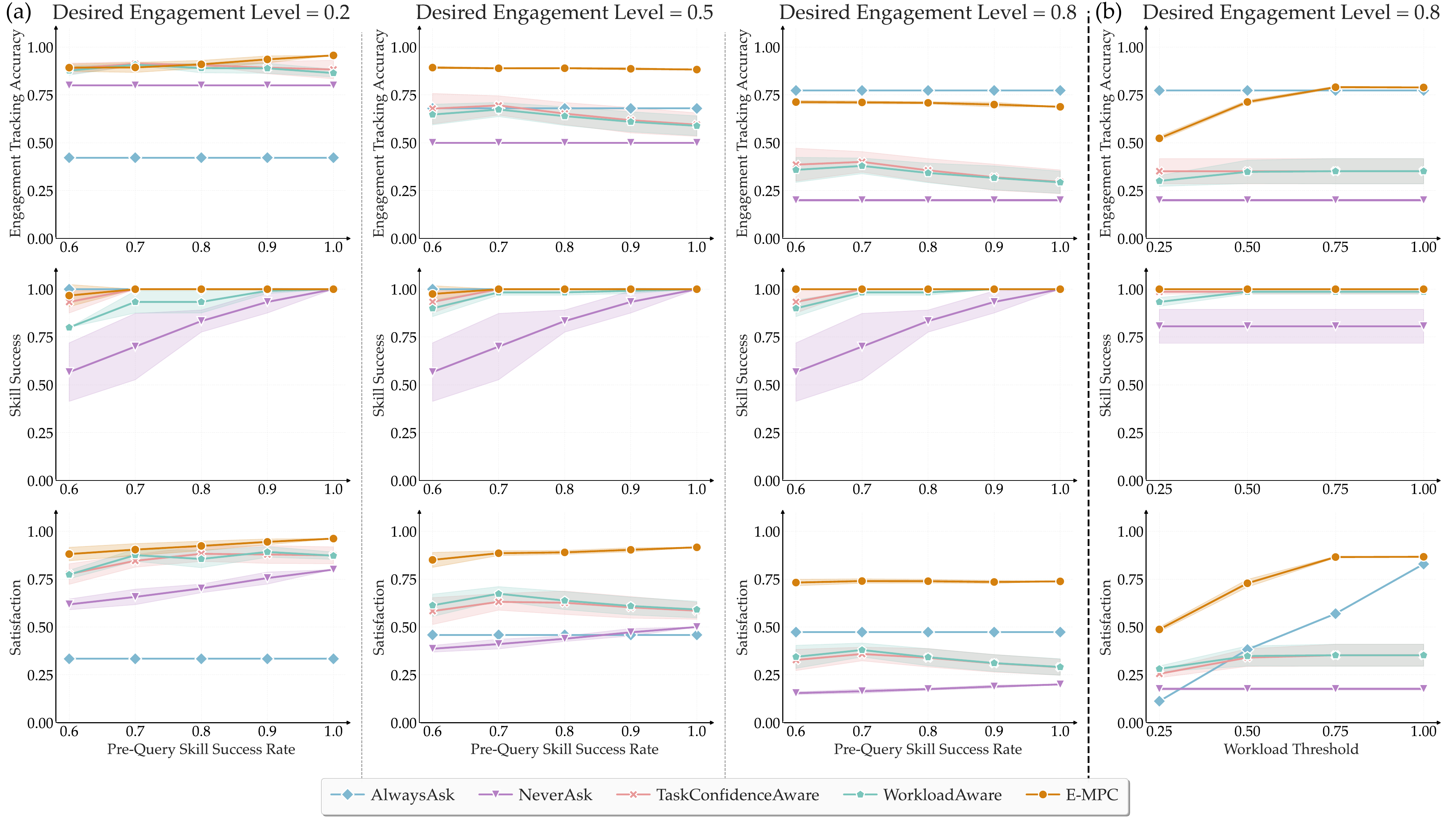}
    \caption{\small{\textbf{Simulation Results}. (a) Performance of methods across varying pre-query skill success rates and user personas. Results are averaged over workload thresholds $\tau_w \in \{0.25, 0.5, 0.75, 1.0\}$. For \textit{WorkloadAware} baseline, we set the value of the hyperparameter $\gamma_{\mathrm{scale}}$ as the value that maximizes the \textit{Satisfaction} metrics in each setting. (b) Performance of methods across varying workload thresholds $\tau_w$ with high-engagement user persona. Results are averaged over pre-query skill success rates $p_{\text{success}} \in \{0.6, 0.7, 0.8, 0.9, 1.0\}$.}} 
    \label{fig:simulation_results}
    \vspace{-20pt}
\end{figure*}

In simulation experiments, we aim to answer the following question: compared to baseline policies, does \textit{E-MPC} optimize user engagement towards a desired target level while achieving task success and maintaining bounded user workload?

\subsection{Baselines} 
We compare \textit{E-MPC} against the following four baselines:
\begin{itemize}
    \item \textit{AlwaysQuery}: asks the \emph{task-assistive} query at every step.
    \item \textit{NeverQuery}: executes all skills autonomously without querying the user.
    \item \textit{TaskConfidenceAware}: asks the \emph{task-assistive} query when the predicted skill success confidence $\mathcal{C}(\phi^i_t, \mathbf{o}_t)$ falls below a threshold. This baseline optimizes solely for task success.
    \item \textit{WorkloadAware}: asks the \emph{task-assistive} query when the skill uncertainty, $1-\mathcal{C}(\phi^i_t, \mathbf{o}_t)$, exceeds a scaled version of the user workload $\gamma_{\mathrm{scale}}w_t$, thereby jointly accounting for task success and user workload. This baseline is the method proposed in \cite{banerjee2025ask}, which is the state-of-the-art method that considers both task success and user workload. 
\end{itemize}

\subsection{Metrics}
We evaluate the performance of different methods using three metrics: \textit{Skill Success}, \textit{Engagement Tracking Accuracy}, and \textit{Satisfaction}. 

\begin{enumerate}

\item \textit{Skill Success} measures the fraction of skill executions in which the skill succeeds within a maximum of $\tau_r$ retries. 

\item \textit{Engagement Tracking Accuracy} evaluates how well a method maintains the desired engagement level $g_{des}$. We define it as one minus the Root-Mean-Square Error (RMSE) between the tracked engagement $g_t$ and $g_{des}$ over a horizon of length $T$: $\mathrm{RMSE} = \sqrt{\frac{1}{T}\sum_{t=1}^{T}\bigl(g_t - g_{des}\bigr)^2}$
, $\mathrm{EngagementAcc} = 1 - \mathrm{RMSE}$.

\item \textit{Satisfaction} captures the overall user experience by jointly accounting for engagement tracking, workload compliance, and task completion. The satisfaction score is defined as
\begin{equation}
\mathrm{Satisfaction} =
\begin{cases}
1 - |g_t - g_{des}|, 
& \begin{aligned}
   &\text{if } w_t \leq \tau_w \text{and skill} \\
   &\text{succeeds in $\tau_r$ retries},\\
  \end{aligned}\\[6pt]
0, 
& \text{otherwise}.
\end{cases}
\end{equation}
This metric reflects that accurate engagement tracking is only meaningful when achieved under bounded workload and reliable task success.
\end{enumerate}
Note that the metrics used in simulation experiments here are not intended to measure whether the engagement model aligns with true user engagement, instead they are used to evaluate whether \textit{E-MPC} can effectively optimize engagement and outperform baseline methods, while assuming our model reflects true user engagement. In Sec.~\ref{sec:user_study}, the user study metrics then evaluate whether our framework with the proposed engagement model aligns with the user experience.
\begin{figure*}
    \centering
    \includegraphics[width=\linewidth]{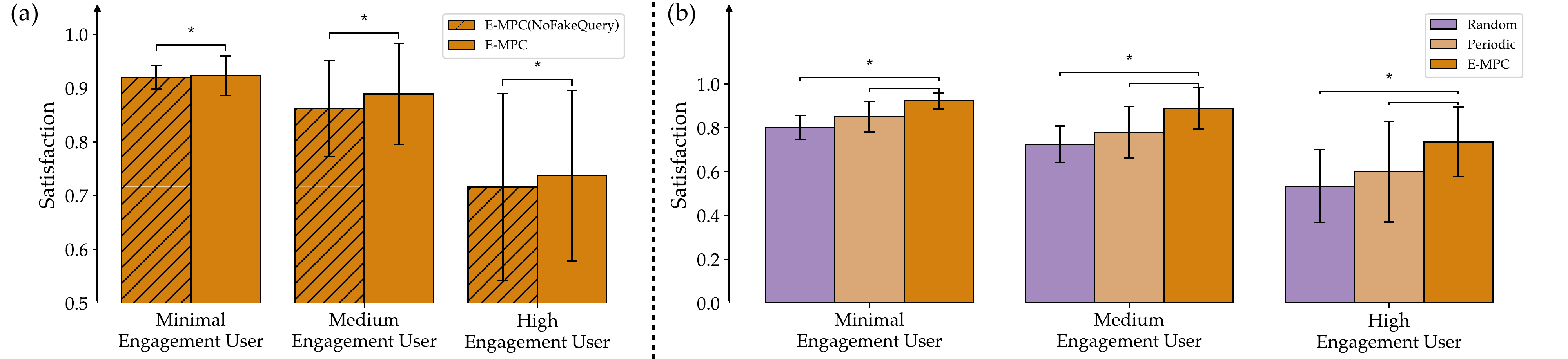}
    \caption{\small{\textbf{Effect of ``Fake" Queries and Heuristic Baselines.} (a) \textit{Satisfaction} for three personas under \textit{E-MPC} and its No-Fake-Queries ablation. (b) \textit{Satisfaction} for three personas under \textit{E-MPC} and two heuristic baselines: \textit{Random} and \textit{Periodic}. Results are averaged over pre-query skill success rates $p_{\text{success}} \in \{0.6, 0.7, 0.8, 0.9, 1.0\}$ and workload thresholds $\tau_w \in \{0.25, 0.5, 0.75, 1.0\}$. $*$ indicates statistical significance $p<0.05$, determined via a Mann-Whitney U test.}}
    \label{fig:fake_queries_and_random_baselines}
    \vspace{-15pt}
\end{figure*}

\subsection{Experiment Settings}
In simulation, the robot can issue two types of queries that are mentioned in Sec.~\ref{sec:query_types}: easy MCQs and hard drawing-based questions. We assign the first query type a query impact $k_{q_{rd}}$ of $0.5$; the second query type a query impact $k_{q_{rd}}$ of $1.0$. Thus, the easy MCQs provide a modest engagement boost with low workload cost, whereas drawing-based queries yield stronger engagement gains but impose substantially higher workload.

The skill confidence $\mathcal{C}(\phi^i_t, \mathbf{o}_t)$ is modeled as a noisy binary estimate, reflecting real-world prediction uncertainty with imperfect true-positive and false-positive rates (TPR$=0.9$, FPR$=0.1$). Each rollout consists of $T=10$ task-planning steps, with a maximum of $\tau_r=2$ retries allowed after a skill execution failure. The MPC prediction horizon is set to $L=5$. 

To capture diverse user engagement preferences, we consider three personas with low, medium, and high desired engagement levels: $g_{\mathrm{des}}\in\{0.2,0.5,0.8\}$. We evaluate each method under varying pre-query skill success rate $p_{\text{success}}=P(\phi^i_t \text{ succeeds without querying})\in\{0.6,0.7,0.8,0.9,1.0\}$ and workload thresholds $\tau_w\in\{0.25,0.5,0.75,1.0\}$. Each condition $(p_{\text{success}}, \tau_w)$ is evaluated over three rollouts.

\subsection{Results}
As shown in Fig.~\ref{fig:simulation_results}(a), \textit{E-MPC} consistently achieves high \textit{Engagement Tracking Accuracy} across all three personas. The \textit{TaskConfidenceAware} and \textit{WorkloadAware} baselines attain comparable tracking performance when the desired engagement level $g_{des}$ is relatively low. However, as $g_{des}$ increases, the advantage of \textit{E-MPC} becomes more pronounced when compared to these two baselines. When $g_{des}$ becomes very high, \textit{AlwaysQuery} can also track engagement well, as it maximizes interaction regardless of workload or task conditions.

In addition, \textit{E-MPC} achieves higher \textit{Skill Success} than all other baselines except \textit{AlwaysQuery}. This improvement arises because \textit{E-MPC} queries the user not only when the predicted skill confidence $\mathcal{C}(\phi^i_t, \mathbf{o}_t)$ is low, but also when interaction is beneficial for maintaining the target engagement level $g_t$. By doing so, \textit{E-MPC} can mitigate the impact of occasional false-positive confidence estimates caused by noisy skill confidences, leading to higher overall success rates.

Finally, \textit{E-MPC} outperforms all baselines on the \textit{Satisfaction} metric. This indicates that \textit{E-MPC} not only tracks engagement effectively, but also ensures task success while keeping user workload bounded. Among the baselines, \textit{WorkloadAware} is the strongest overall; however, because it does not explicitly account for engagement, its satisfaction scores drop substantially for medium- and high-engagement personas. \textit{AlwaysQuery} achieves strong engagement tracking and high task success for highly engaged users, but its disregard for workload leads to reduced satisfaction.

\subsection{Workload Thresholds}
Fig.~\ref{fig:simulation_results}(b) compares baseline methods and \textit{E-MPC} under a high desired engagement level ($g_{des}=0.8$). When the maximum allowable workload threshold $\tau_w$ is low, \textit{E-MPC} tracks engagement less accurately than \textit{AlwaysAsk}, since it prioritizes maintaining workload compliance over engagement interaction. Importantly, this prioritization leads to a better overall user experience: this is reflected in the \textit{Satisfaction} metric, where \textit{E-MPC} achieves substantially higher satisfaction for low to medium workload thresholds.

\subsection{Ablations}
In addition to the baselines above, we also compare \textit{E-MPC} against the following ablations: (1) \textit{NoWorkloadCst}: our method without the user workload constraint; (2) \textit{NoEngagementTrack}: our method without the engagement tracking cost. 

In Table \ref{table:ablations}, beyond \textit{Engagement Tracking Accuracy} and \textit{Satisfaction}, we further report the \textit{Workload Compliance Rate}, which measures the fraction of timesteps during which the user workload remains below the maximum allowable threshold $\tau_w$. When $g_{des}$ is low or moderate, \textit{NoWorkloadCst} performs comparably to \textit{E-MPC}. However, as $g_{des}$ increases to $0.8$, \textit{NoWorkloadCst} frequently violates the workload constraint, exceeding $\tau_w$ in roughly $50\%$ of timesteps. Since \textit{NoEngagementTrack} does not account for engagement, it consistently underperforms \textit{E-MPC} across all settings.

\begin{table}[!t]
\centering
\scriptsize
\setlength{\tabcolsep}{4pt}
\renewcommand{\arraystretch}{0.9}
\caption{Performance of \textit{E-MPC} and its ablations across three desired engagement levels (pre-query skill success rate $p_{\text{success}}=80\%$ and workload threshold $\tau_w=0.5$)}
\vspace{-5pt}
\label{table:ablations}
\begin{tabular}{c|p{0.8 in}|c|c|c}
\toprule
\makecell{$g_{des}$} & Mode & \makecell{Engagement\\Acc.}$\uparrow$ & \makecell{Workload\\Compliance (\%)}$\uparrow$ & Satisfaction$\uparrow$ \\ 
\midrule
\multirow{3}{*}{${0.2}$}
& E-MPC
& $0.910\!\pm\!0.021$
& $100.0\!\pm\!0.000$
& $\mathbf{0.933\!\pm\!0.012}$\\

& NoWorkloadCst
& $0.910\!\pm\!0.021$
& $100.0\!\pm\!0.000$
& $\mathbf{0.933\!\pm\!0.012}$\\

& NoEngagementTrack
& $0.862\!\pm\!0.013$
& $100.0\!\pm\!0.000$
& $0.880\!\pm\!0.013$\\

\midrule
\multirow{3}{*}{${0.5}$}
& E-MPC
& $0.917\!\pm\!0.003$
& $100.0\!\pm\!0.000$
& $\mathbf{0.942\!\pm\!0.007}$\\

& NoWorkloadCst
& $0.919\!\pm\!0.001$
& $97.62\!\pm\!4.124$
& $0.922\!\pm\!0.035$\\

& NoEngagementTrack
& $0.594\!\pm\!0.025$
& $100.0\!\pm\!0.000$
& $0.607\!\pm\!0.024$\\

\midrule
\multirow{3}{*}{${0.8}$}
& E-MPC
& $0.720\!\pm\!0.013$
& $97.22\!\pm\!4.811$
& $\mathbf{0.730\!\pm\!0.028}$\\

& NoWorkloadCst
& $0.786\!\pm\!0.000$
& $50.00\!\pm\!0.000$
& $0.382\!\pm\!0.000$\\

& NoEngagementTrack
& $0.300\!\pm\!0.027$
& $100.0\!\pm\!0.000$
& $0.307\!\pm\!0.024$\\

\bottomrule
\end{tabular}
\end{table}

\subsection{Effect of ``Fake" Queries}
We also explore the effect of engagement-oriented (``fake'') queries in simulation. In our setting, these queries take the form of easy MCQs that are intended solely to increase user engagement and do not help the robot complete the task. We compare \textit{E-MPC} against an ablation that excludes these engagement-oriented queries and permits only \emph{task-assistive} queries. As shown in Fig.~\ref{fig:fake_queries_and_random_baselines}(a), incorporating engagement-oriented queries yields a statistically significant improvement in the \textit{Satisfaction} metric across all three personas. This improvement arises because these ``fake" queries provide engagement boosts while imposing relatively low additional workload, enabling the robot to better align with user engagement preferences without exceeding maximum allowed workload.

\subsection{Comparison with Heuristic Baselines}
We compare E-MPC against two heuristic baselines that approximate engagement via simple query scheduling. The \textit{Random} baseline queries with probability $P(\mathbf{q}_t \neq \mathbf{0}) = g_{\mathrm{des}}$, while the \textit{Periodic} baseline queries at fixed intervals determined by $g_{\mathrm{des}}$. As shown in Fig.~\ref{fig:fake_queries_and_random_baselines}(b), \textit{E-MPC} consistently achieves higher satisfaction across all user types compared to heuristic baselines. While \textit{Random} and \textit{Periodic} regulate query frequency, they fail to adapt interaction to task uncertainty or user state, leading to suboptimal behavior. In contrast, \textit{E-MPC} explicitly models engagement and selects both when and how to query, leading to better engagement tracking and higher satisfaction.

%% file: 08_user_study.tex
\label{sec:user_study}
\begin{figure*}[!t]
    \centering
    \includegraphics[width=\linewidth]{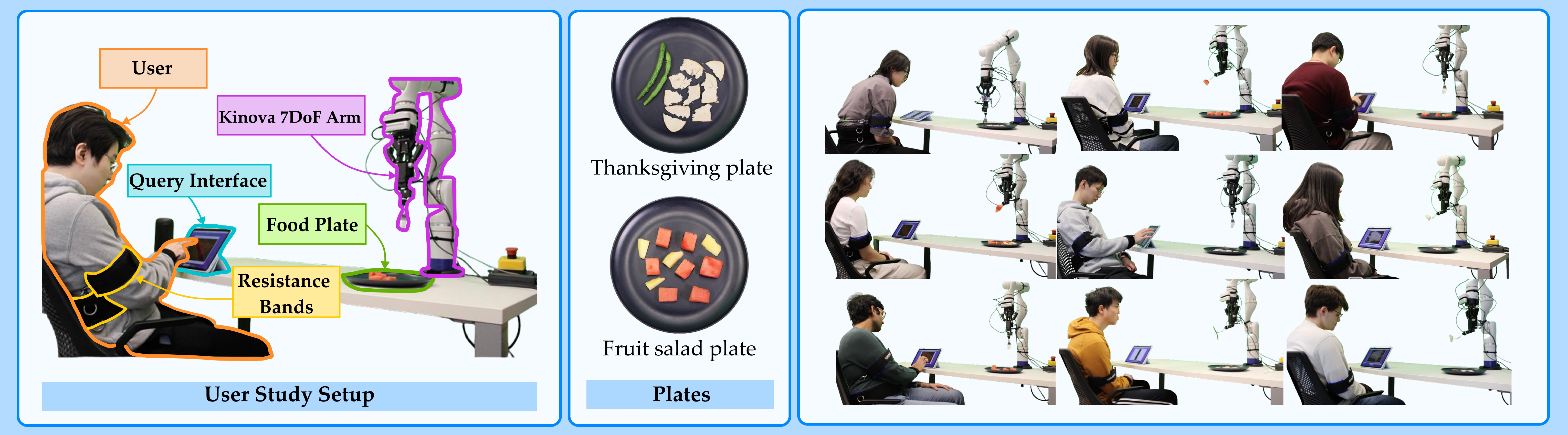}
    \caption{\small{\textbf{Study Setup}. (left) User study setup. (middle) Plates used in the study: a Thanksgiving plate and a fruit salad plate. (right) Participants in the user study with emulated mobility limitations using resistance bands (faces shown with full permission).}}
    \vspace{-10pt}
    \label{fig:user_study_setup}
\end{figure*}
\begin{figure*}[!t]
    \centering
    \includegraphics[width=\linewidth]{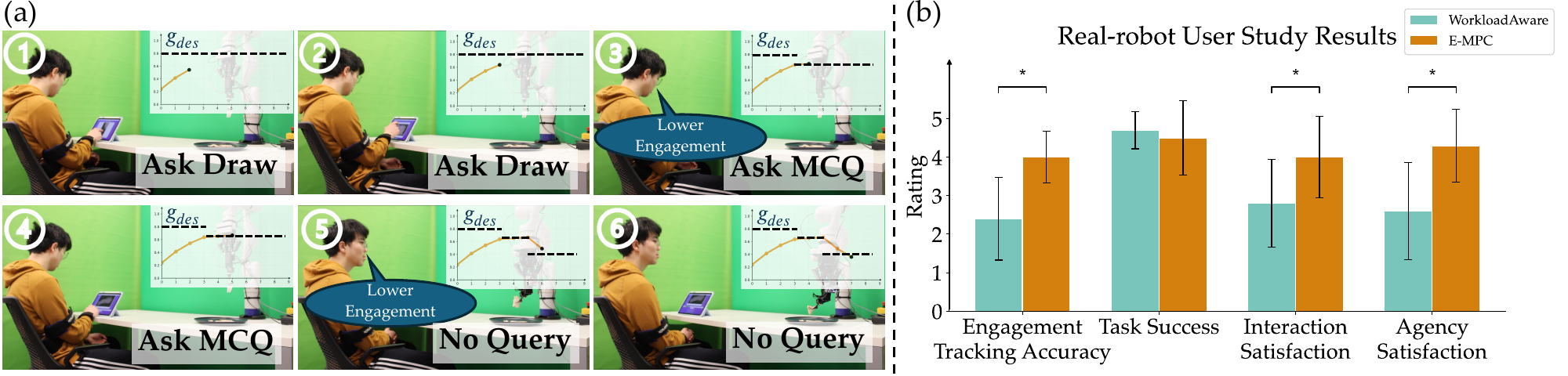}
    \caption{\small{\textbf{User Study Results}. (a) Visualization of a rollout from the user study. The user initially set the $g_{des}=0.8$. During the process, the user lowered $g_{des}$ twice. \textit{E-MPC} adapts to the updated $g_{des}$. (b) Participants ratings on four subjective metrics. * indicates statistical significance ($p<0.05$), determined via a Mann-Whitney U test.}}
    \label{fig:user_study}
    \vspace{-20pt}
\end{figure*}
We conducted a Cornell University IRB-approved real-robot user study on a robot-assisted bite acquisition system to evaluate the effectiveness of \textit{E-MPC}. The study involved 10 participants (4 female, 6 male; ages 19-29). Participants provided informed consent and were compensated for their time. 

A Kinova Gen3 7-DoF arm with a Robotiq 2F-85 gripper is used in the study (Fig.~\ref{fig:user_study_setup}(left)). The tool attached to the end-effector is adapted from pre-existing tools \cite{jenamani2024flair}. As described in Sec.~\ref{sec:query_types}, the robot-assisted bite acquisition system supports two types of user queries. Among them, only the drawing-based query is needed for successful task completion when the robot becomes uncertain. The multiple-choice question about food shape is a ``fake" query that is related to the task but does not help the execution of the skill. During each study session, the robot attempts to acquire 10 food items from the plate, and up to $\tau_r = 2$ retries are permitted. 

\subsection{Baseline and Metrics}
Since \textit{WorkloadAware} is the strongest baseline overall based on our simulation results and is the state-of-the-art method proposed in \cite{banerjee2025ask}, we therefore select \textit{WorkloadAware} as the primary comparison method alongside \textit{E-MPC} in the real-robot user study. 

The user study metrics are designed to evaluate whether our framework with the proposed engagement model aligns with what the users care about. Therefore, these metrics are directly self-reported by participants based on their responses to the following five questions, which we use to evaluate user experience and system performance. All user study metrics except \textit{Workload Compliance} are reported on a 5-point Likert scale, while \textit{Workload Compliance} is reported as a binary (yes/no) outcome:

\begin{enumerate}    
    \item \textit{Engagement Tracking Accuracy}: {\fontfamily{qcr}\selectfont \emph{"How successful was it for keeping you engaged at your desired engagement level?"}}   
    \item \textit{Workload Compliance}: {\fontfamily{qcr}\selectfont \emph{"Is the workload you felt during the task acceptable?"}}    
    \item \textit{Task Success}: {\fontfamily{qcr}\selectfont\emph{"How satisfied are you with how successfully the robot completed the task?"}}
    \item \textit{Interaction Satisfaction}: {\fontfamily{qcr}\selectfont\emph{"How satisfied are you with how the robot interacted with you?"}}
    \item \textit{Agency Satisfaction}: {\fontfamily{qcr}\selectfont\emph{"How satisfied are you with how much control/agency you felt during the task?"}}
    \end{enumerate}

\subsection{Study Procedure}
Before the user study, we perform several user-specific calibration steps. First, we calibrate the query impact parameter $k_{q_{rd}}$. Specifically, we present each of the two feasible query types to the participant and ask them to rate the maximum engagement level they would expect if that query were issued at every timestep over a prolonged interaction. Next, we collect the participant's desired engagement level $g_{des}$ using a continuous slider, where $0$ indicates no engagement and $1$ indicates full engagement. Finally, the participant specifies the maximum workload they are willing to tolerate during the task via a slider, which we set as the workload threshold $\tau_w$. After calibration, each subject evaluates the baseline method and \textit{E-MPC} on the same plate. Across the study, we use two plates, a Thanksgiving plate and a fruit salad plate, with plate assignment counterbalanced between participants (Fig.~\ref{fig:user_study_setup} (middle)). 

Following each method, participants completed a post-method questionnaire with five questions corresponding to our five evaluation metrics. After completing the full study, participants completed a post-study questionnaire in which they selected their preferred method and provided additional qualitative feedback. 

The calibration details and post-method questionnaire are described in the Appendix.

\subsection{Results}
From user-reported ratings shown in Fig.~\ref{fig:user_study}(b) (see Appendix for detailed statistics), we find that \textit{E-MPC} achieves significantly better performance than the baseline on \textit{Engagement Tracking Accuracy}, \textit{Interaction Satisfaction}, and \textit{Agency Satisfaction}. Both the \textit{WorkloadAware} \cite{banerjee2025ask} and \textit{E-MPC} perform comparably on the \textit{Task Success} metric. In addition, all 10 participants reported that \textit{Workload Compliance} is achieved, meaning that their workload remained acceptable under both methods throughout the study. In the post-study questionnaire, 9 out of 10 participants preferred \textit{E-MPC} over the state-of-the-art baseline \cite{banerjee2025ask} for providing a more aligned engagement experience. One participant noted, ``\textit{I do think more control is better for me before it gets tiring},'' and another commented, ``\textit{The balance of engagement and successful runs brought satisfaction}.'' This feedback reinforces our quantitative findings, suggesting that participants valued the increased engagement enabled by \textit{E-MPC} while maintaining task success and a comfortable interaction experience. 

Notably, among all queries that the robot asked users during the user study, 51.65\% were ``fake" queries, while 48.35\% were \emph{task-assistive} queries. These results suggest that \textit{E-MPC} effectively maintains user engagement near their desired level with strategic use of ``fake" queries. 

To complement self-reports, we also include an additional eye-gaze analysis from study videos. We observe that when user engagement $g_t$ deviates from the desired level $g_{des}$, users' gaze behavior becomes less task-focused, suggesting a measurable behavioral difference associated with engagement misalignment, which is consistent with prior work \cite{buono2023assessing, rich2010recognizing}. Detailed analysis and results are provided in the Appendix.

%% file: 09_discussions.tex
We present \textit{E-MPC}, a human-in-the-loop framework for robotic systems that explicitly models and optimizes user engagement. \textit{E-MPC} enables robots to reason about when and how to involve users. Through simulation across diverse user personas and a real-robot bite-acquisition user study, we demonstrated that \textit{E-MPC} improves interaction satisfaction and perceived agency without sacrificing task success or exceeding user workload limits. These results highlight the importance of user-engagement-aware interaction planning for human-in-the-loop policies, and open new opportunities for designing robotic systems that better support user-centered assistance.

An important direction for future work is to make engagement regulation more transparent and negotiable between the user and the robot. In the current formulation, \textit{E-MPC} can use ``fake'' queries that do not directly affect task execution to increase engagement with relatively low additional workload. This design helps separate engagement regulation from task assistance, but it may also raise concerns about transparency and deception: users may interpret these queries as task-relevant even when they are not. An alternative approach would be to increase engagement through explicit and transparent interaction strategies.

Our findings also motivate a deeper investigation into the relationship between user engagement, workload, task success, and perceived agency. In our current framework, engagement and workload are modeled explicitly, whereas perceived agency is evaluated only through self-report. Although engagement may contribute to agency, agency is a distinct construct that cannot be fully explained by engagement alone. For example, maintaining engagement near a desired level may help users feel that they can meaningfully influence the robot’s behavior, but this effect may depend on whether the interaction also feels manageable and whether the robot succeeds at the task. Excessive workload may make users feel burdened rather than agentic, while too little involvement may make them feel passive or disconnected from the task outcome. Moreover, perceived agency may depend on other factors beyond engagement, workload, and task success, such as interface design, transparency, and user expectations. Therefore, modeling perceived agency would require a dedicated study, which is beyond the scope of this project.

One limitation of our work is the reliance on an explicit calibration procedure. In the current system, the pre-task calibration process is used both to set the user’s desired engagement level and workload threshold, and to calibrate the engagement model using hypothetical interaction scenarios, which can be noisy. Future work could improve this process by estimating both the user's current engagement level and desired engagement level online from behavioral signals, interaction history, and direct user feedback.

A second limitation is that our current formulation models interaction primarily through query actions. Queries provide a convenient and interpretable mechanism for eliciting user input. However, real-world human-robot interaction supports a broader range of interaction modalities, including shared control, conversational feedback, implicit corrections, physical guidance, explanations, and socially grounded behaviors, etc. Extending \textit{E-MPC} beyond explicit queries would allow the framework to capture richer and more natural forms of engagement.

Finally, while our user study evaluates \textit{E-MPC} on a real robot in an assistive bite-acquisition task, it does not include participants with mobility limitations, who are a key target population for assistive robotics. Evaluating the framework with users who have lived experience with mobility constraints is an important next step. Such studies may reveal different engagement preferences, workload sensitivities, and would provide a stronger evaluation of whether \textit{E-MPC} can improve user engagement during real-world assistive robot deployments.

%% file: 10_appendix.tex
\begin{appendices}
\section*{Appendix}
\setcounter{figure}{0}
\renewcommand{\thefigure}{A\arabic{figure}}
\setcounter{table}{0}
\renewcommand{\thetable}{A\arabic{table}}
\label{appendix}

\subsection{Simulation results Across Workload Thresholds}
\begin{figure*}[t]
    \centering
    \includegraphics[width=\linewidth]{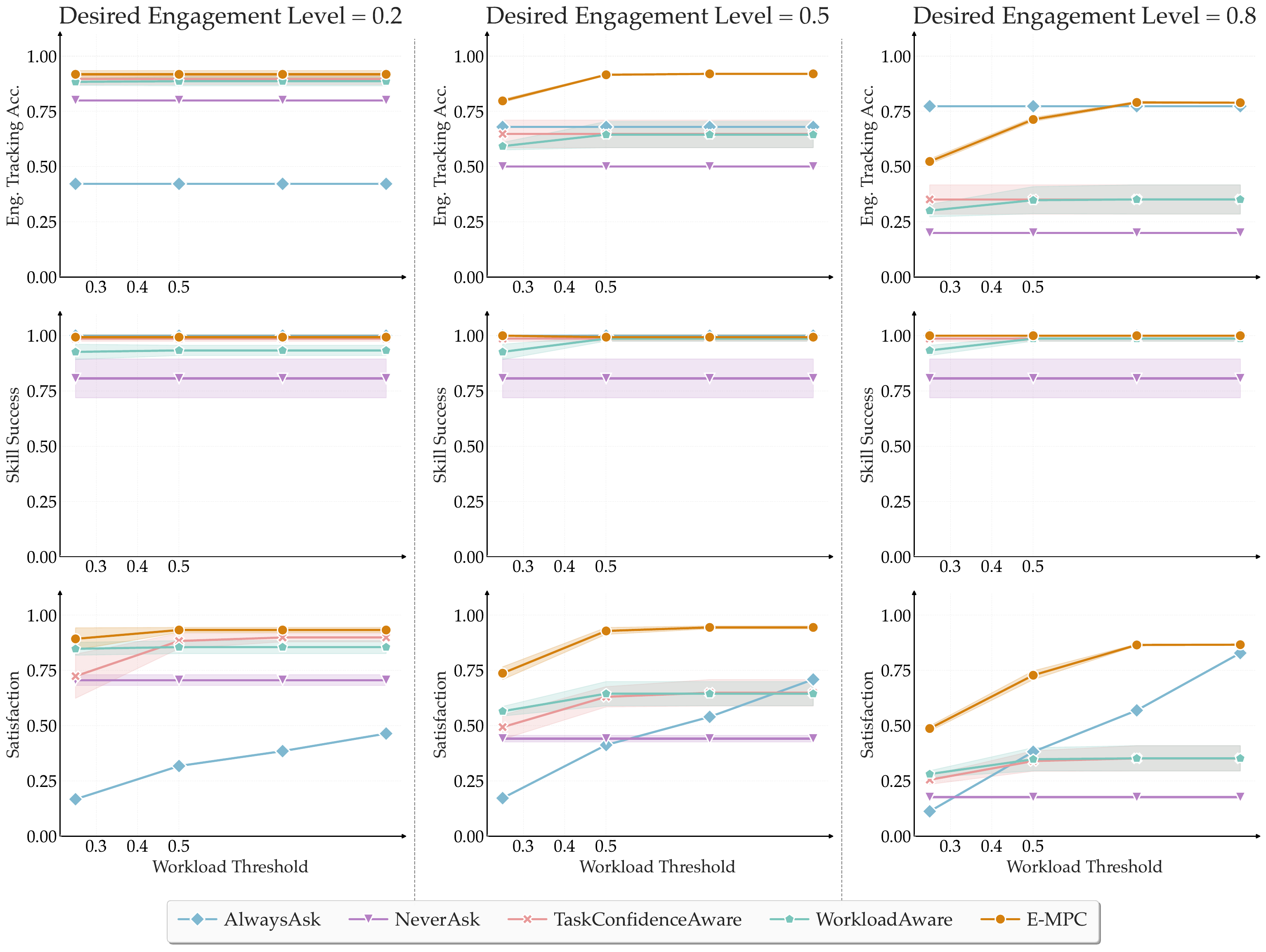}
    \caption{Performance of methods across varying workload thresholds and user personas. Results are averaged over pre-query skill success rate $p_{\text{success}} \in \{0.6, 0.7, 0.8, 0.9, 1.0\}$. For \textit{WorkloadAware} baseline, we set the value of the hyperparameter $\gamma_{\mathrm{scale}}$ as the value that maximizes the \textit{Satisfaction} metrics in each setting.}
    \label{fig:appendix_workload_sim}
\end{figure*}

Fig.~\ref{fig:appendix_workload_sim} shows performance across workload thresholds and user personas. \textit{E-MPC} achieves high \textit{Engagement Tracking Accuracy} and \textit{Satisfaction} across desired engagement levels. As $\tau_w$ increases, \textit{AlwaysAsk} improves because frequent querying incurs less workload penalty; however, \textit{E-MPC} consistently achieves the highest \textit{Satisfaction}.

\subsection{Effect of Different Query Impacts $K_{q_{rd}}$} 
\begin{figure*}[b]
    \centering
    \vspace{-5pt}
    \includegraphics[width=\linewidth]{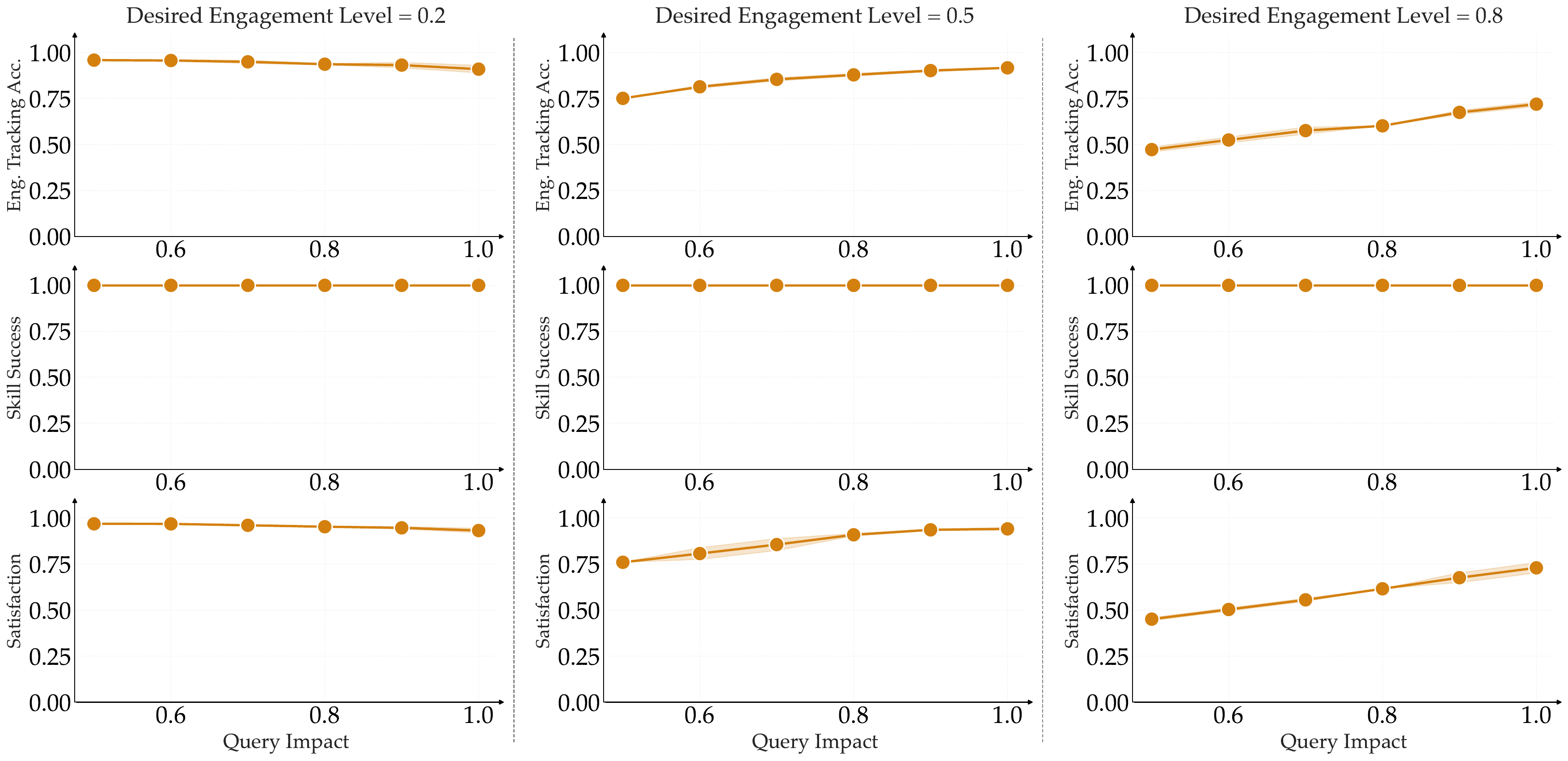}
    \caption{Performance \textit{E-MPC} across varying query impacts $K_{q_{rd}}$ and user personas. The values on the x-axis are $K_{q_{rd}}$ of the hard drawing-based questions, while the $K_{q_{rd}}$ for easy multiple-choice questions is set to half of that value. Results are reported with $80\%$ pre-query skill success rate and $0.5$ as the workload threshold.}
    \label{fig:appendix_query_impact_sim}
\end{figure*}

Fig.~\ref{fig:appendix_query_impact_sim} evaluates \textit{E-MPC} under different query impact values. We vary $K_{q_{rd}}$ for hard drawing-based queries and set the impact of easy multiple-choice queries to half that value. As $g_{des}$ increases, queries with lower $K_{q_{rd}}$ increasingly struggle to track the target engagement. This behavior arises because the maximum engagement that can be induced by a query is bounded by its query impact $K_{q_{rd}}$.

\subsection{Effect of Different $\lambda_g$ Values in Interaction Dynamics Model}
\begin{figure*}[!t]
    \centering
    \includegraphics[width=\linewidth]{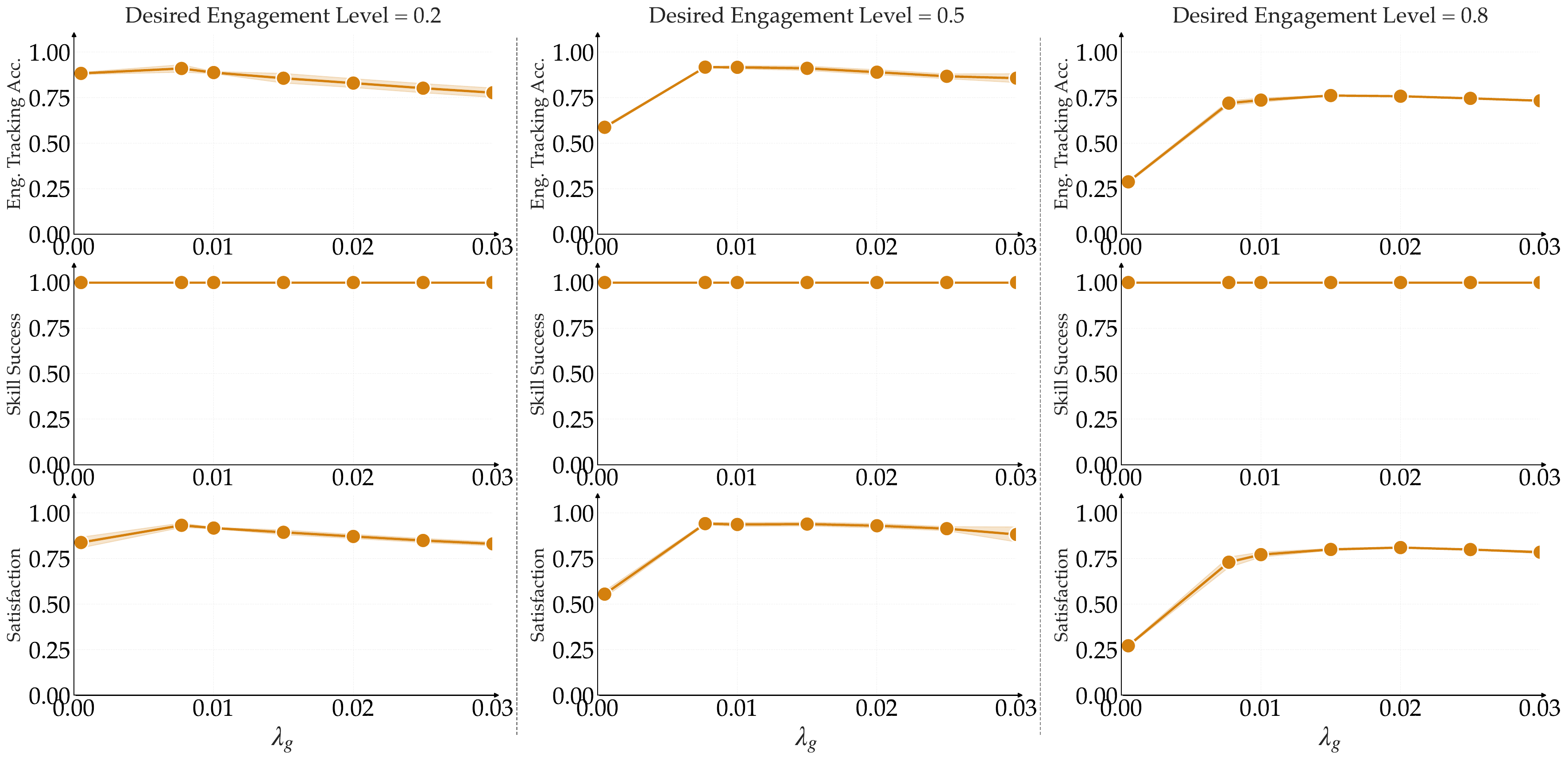}
    \caption{Performance \textit{E-MPC} across varying values of $\lambda_g$ and user personas. Results are reported with $80\%$ pre-query skill success rate and $0.5$ as the workload threshold.}
    \label{fig:appendix_lambda_g_sim}
\end{figure*}

Fig.~\ref{fig:appendix_lambda_g_sim} evaluates different values of $\lambda_g$. A larger $\lambda_g$ corresponds to less persistent engagement memory and greater sensitivity to the instantaneous engagement impulse, whereas a smaller $\lambda_g$ implies more persistent engagement memory and reduced sensitivity to engagement changes induced by queries. When $\lambda_g$ is too small, \textit{E-MPC} struggles to raise engagement for high $g_{des}$; when it is too large, single queries can cause overshoot, especially for low $g_{des}$.

\subsection{Skill Confidence in the Bite Acquisition System}
The confidence score for each acquisition skill is modeled as a binary value from two sources: (1) the uncertainty of the transformer-based skill parameter estimation model, and (2) a failure identification module. We obtain the uncertainty of the transformer-based skill parameter estimation model using Monte Carlo Dropout, which introduces stochasticity in the model. Specifically, following the procedure in \cite{banerjee2026modularhil}, we activate dropout at inference time and perform 16 stochastic forward passes. This yields a distribution of predictions, from which we compute the output variance as a measure of model confidence. The failure identification module for skewering and scooping checks if the predicted skewering/scooping point is within the SAM3 \cite{carion2025sam3segmentconcepts} segmentation mask of the food item. There is no failure identification module for twirling skill.

\subsection{Adapting \textit{E-MPC} to Other Robotic Systems}
Beyond bite acquisition in robot-assisted feeding, \textit{E-MPC} readily extends to other physical human-robot interaction scenarios that involve long-horizon decision-making. For example, in robot-assisted hair grooming, \textit{E-MPC} can reason over grooming skill selection (e.g., combing or spraying) and selectively query users to balance engagement, workload, and task progress. In robot-assisted meal preparation, \textit{E-MPC} can coordinate sequential manipulation skills (e.g., cutting, stirring, plating) while adapting interaction pattern to user preferences. Similarly, in human–robot collaborative assembly, \textit{E-MPC} can modulate when to request user input for alignment, verification, or correction, maintaining user engagement. Across these domains, only the skill library, query set, and task-specific confidence models need to be redefined, while the core engagement-aware MPC formulation remains unchanged.

\subsection{Calibration Questions Asked in User Study}
\begin{lstlisting}[style=constraintprompt, label={lst:constraint_prompt}]
1. [Show a demo of the hard drawing-based question] How engaged would you feel if you keep interacting with the robot by answering this type of drawing-based questions?
2. [Show a demo of the easy MCQ] How engaged would you feel if you keep interacting with the robot by answering this type of food shape questions?
3. How engaged do you want to be in this task? (Note: Your answer would change how often you are asked by the robot and the type of questions)
4. What is the largest workload you can accept in this task? (Note: 10% roughly corresponds to asking you one "draw pick up points" question in 5 minutes, 100% roughly corresponds to asking you "draw pick up points" questions 10 times in 5 minutes.)
\end{lstlisting}

\subsection{Post-Method Questions Asked in User Study}
\begin{lstlisting}[style=constraintprompt, label={lst:constraint_prompt}]
1. For the last method, how successful was it for keeping you engaged at your desired engagement level?
2. For the last method, is the workload you felt during the task acceptable?
3. For the last method, how satisfied are you with how successfully the robot completed the task?
4. For the last method, how satisfied are you with how the robot interacted with you?
5. For the last method, how satisfied are you with how in control/agency you felt during the task?
\end{lstlisting}

\subsection{Statistics of User Study}
Table~\ref{table:user_study_stats} shows detailed statistics of the user study results.

\subsection{Eye-gaze Analysis}
The additional eye-gaze analysis is conducted using study videos by estimating head pose (using head direction as a proxy for gaze) and projecting it onto task-relevant objects (robot, food plate, tablet). From Table~\ref{table:gaze_hit}, we observe that when user engagement $g_t$ deviates substantially from the desired level $g_{des}$ (by more than $\pm 0.3$), gaze behavior becomes less task-focused. Specifically, gaze misses all target objects at a rate 123\% higher than when engagement is near $g_{des}$, while gaze directed at the food plate (the shared object of interest) drops to 37\% of its baseline rate. This difference is highly significant ($p < 0.001$, $\chi^2$ test). These results suggest that misalignment between actual and desired engagement is associated with reduced attention to task-relevant objects. Consistent with this observation, head motion is also more variable under engagement deviation, with a yaw standard deviation of $41.17^\circ$ compared to $21.71^\circ$ when engagement is close to $g_{des}$, as shown in Table~\ref{table:head_pose}. 

\begin{table}[b]
\centering
\caption{Subjective user study results. Values are reported as mean $\pm$ standard deviation.}
\label{tab:user_study_subjective}
\resizebox{\columnwidth}{!}{
\begin{tabular}{lccc}
\toprule
Metric & WorkloadAware & E-MPC & $p$-value \\
\midrule
Engagement Tracking Accuracy & $2.40 \pm 1.07$ & $\mathbf{4.00 \pm 0.67}$ & $0.003$ \\
Task Success & $\mathbf{4.70 \pm 0.48}$ & $4.50 \pm 0.97$ & $0.925$ \\
Interaction Satisfaction & $2.80 \pm 1.14$ & $\mathbf{4.00 \pm 1.05}$ & $0.032$ \\
Agency Satisfaction & $2.60 \pm 1.26$ & $\mathbf{4.30 \pm 0.95}$ & $0.007$ \\
\bottomrule
\end{tabular}
}
\label{table:user_study_stats}
\end{table}

\begin{table}[b]
\centering
\caption{Gaze hit rates under Close to $g_{des}$ vs. Deviated Engagement.}
\begin{tabular}{lcc}
\toprule
Category & Close to $g_{des} (\%)$ & Deviated (\%) \\
\midrule
Robot       & 64.43 & 73.35 \\
Food plate  & 11.51 & 4.33 \\
iPad        & 9.27 & 4.14 \\
No hit      & 14.79 & 18.18 \\
\bottomrule
\end{tabular}
\label{table:gaze_hit}
\end{table}

\begin{table}[b]
\centering
\caption{Head pose statistics under Close to $g_{des}$ vs. Deviated Engagement.}
\begin{tabular}{lcccc}
\toprule
 & \multicolumn{2}{c}{Close to $g_{des}$} & \multicolumn{2}{c}{Deviated} \\
\cmidrule(lr){2-3} \cmidrule(lr){4-5}
Metric & Mean ($^\circ$) & Std ($^\circ$) & Mean ($^\circ$) & Std ($^\circ$) \\
\midrule
Pitch (deg) & -38.28 & 28.06 & -30.57 & 27.68 \\
Yaw (deg)   & -58.17 & 21.71 & -53.25 & 41.17 \\
Roll (deg)  &  41.18 & 30.39 &  32.60 & 32.50 \\
\bottomrule
\end{tabular}
\label{table:head_pose}
\end{table}

\end{appendices}